\documentclass{article}
\usepackage{ijcai23}

\usepackage{times}
\usepackage{soul}
\usepackage{url}
\usepackage[hidelinks]{hyperref}
\usepackage[utf8]{inputenc}
\usepackage[small]{caption}
\usepackage{graphicx}
\usepackage{amsmath}
\usepackage{amsthm}
\usepackage{amsfonts}
\usepackage{booktabs}
\usepackage{algorithm}
\usepackage[switch]{lineno}
\usepackage{mathrsfs}
\usepackage{subcaption}
\usepackage{amsmath,amssymb}
\usepackage{tikz}
\usepackage{chngcntr}
\usepackage{algpseudocode}
\usepackage{threeparttable}
\usepackage{xpatch}
\xpretocmd{\part}{\setcounter{section}{0}}{}{}

\newcommand{\argmax}{\mathop{\rm argmax}\limits}

\title{TNF: Tri-branch Neural Fusion for Multimodal Medical Data Classification}

\author{
Tong Zheng$^{1,2}$
\and
Shusaku Sone$^{1}$\and
Yoshitaka Ushiku$^{1}$\and
Yuki Oba$^{1,3}$\and
Jiaxin Ma$^{1}$
\affiliations
$^1$OMRON SINIC X Corporation\\
$^2$Nagoya University\\
$^3$Tsukuba University\\
}

\begin{document}

\maketitle

\begin{abstract}
    This paper presents a Tri-branch Neural Fusion (TNF) approach designed for classifying multimodal medical images and tabular data. It also introduces two solutions to address the challenge of label inconsistency in multimodal classification. Traditional methods in multi-modality medical data classification often rely on single-label approaches, typically merging features from two distinct input modalities. This becomes problematic when features are mutually exclusive or labels differ across modalities, leading to reduced accuracy. To overcome this, our TNF approach implements a tri-branch framework that manages three separate outputs: one for image modality, another for tabular modality, and a third hybrid output that fuses both image and tabular data. The final decision is made through an ensemble method that integrates likelihoods from all three branches. We validate the effectiveness of TNF through extensive experiments, which illustrate its superiority over traditional fusion and ensemble methods in various convolutional neural networks and transformer-based architectures across multiple datasets.
\end{abstract}

\section{Introduction}

Over the past few decades, medical data classification has been widely studied and become a critical part in medical data processing. Medical data classification can help accurately determine the location of patients' lesions and reduce the workload of doctors in the treatment process~\cite{Wang2022Medical}. Most medical data classification models are created for single modality such as image data~\cite{singh20203d}, video data~\cite{funke2019video}, and text data~\cite{pagad2022clinical}. On the other hand, multimodal learning achieved substantial progress in areas such as vision and language learning~\cite{wang2022multimodal,Shagun2021Multimodal}, video learning~\cite{panda2021adamml} and autonomous driving~\cite{xiao2020multimodal}. Multimodal learning was also applied in biomedical data~\cite{acosta2022multimodal}. Consequently, applying multimodal learning in multi-modality medical data analysis is the recent trend~\cite{amal2022use,shaik2023survey,taleb2021multimodal,yao2017deep}.

\begin{figure}[H]
  \centering
  \includegraphics[width=0.49\textwidth]{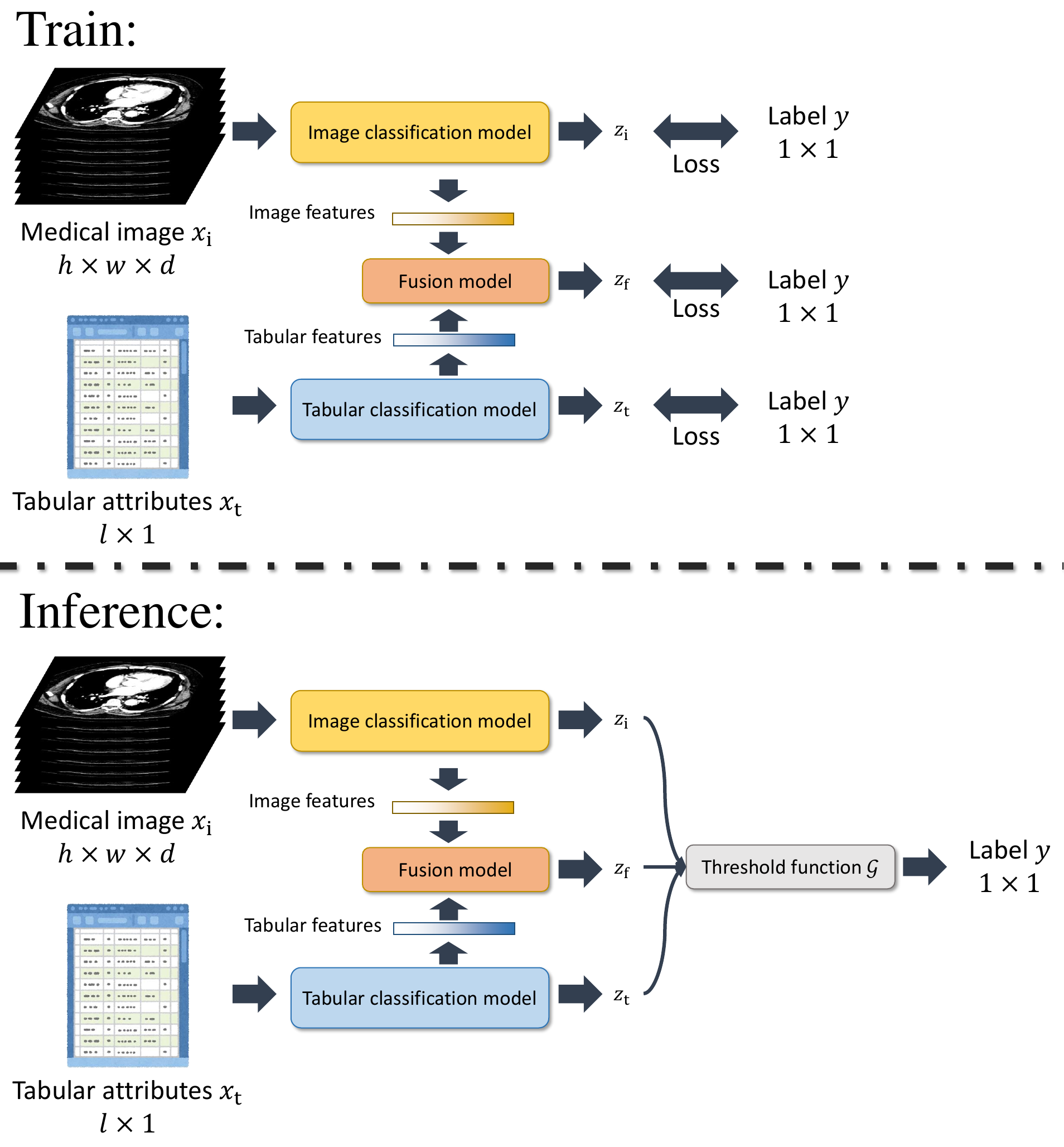}
  \caption{TNF's overall structure. The input are medical image $\boldsymbol{x}_{\rm{i}}$ and corresponding tabular attributes $\boldsymbol{x}_{\rm{t}}$. $\boldsymbol{x}_{\rm{i}}$ and $\boldsymbol{x}_{\rm{t}}$ are inputted into image classification model and tabular classification model respectively to obtain likelihood $\boldsymbol{z}_{\rm{i}}$ and $\boldsymbol{z}_{\rm{t}}$. Image features and tabular features are inputted into fusion model to get likelihood $\boldsymbol{z}_{\rm{f}}$. $\boldsymbol{z}_{\rm{i}}$, $\boldsymbol{z}_{\rm{t}}$ and $\boldsymbol{z}_{\rm{f}}$ are processed by threshold function $\mathcal{G}$ to get the classification result $\hat{y}$. }
  \label{Totalnet}
\end{figure}

In diagnosing certain diseases such as Alzheimer's, clinicians use diagnostic tools combined with medical history and other information, including neurological exams, cognitive and functional assessments, medical imaging (MRI, CT, Ultrasound) and cerebrospinal fluid or blood tests to make an accurate diagnosis~\cite{AlzheimersAssoc2023}. The aforementioned process will generate two modalities of medical data: one is medical image data including MRI, CT or Ultrasound images; another is medical tabular data including patient information such as gender, age and diagnostic results. For automatically diagnosing based on the medical image and tabular data, a multimodal medical data classification approach is desired. The classification method takes medical image and tabular data as input, and gives a diagnostic result as output.

In multimodal classification, there are conventionally two distinct approaches. The first is ensemble~\cite{rokach2010pattern,kumar2016ensemble}, which trains multiple models for different modalities, and ensembles the outputs of multiple models in the inference stage to obtain the final classification result. The second method is fusion, which extracts features of different modalities, then fuses the features and outputs the final classification result~\cite{zhang2021information}. Ensemble has higher robustness, but inferior in accuracy and interpretability than the fusion ones. Fusion has potentially higher accuracy, but easier to suffer from overfitting, and has lower flexibility (require all modalities to be available at both training and inference times). A method that can overcome the respective shortcomings of ensemble and fusion needs to be proposed.

In this paper, we propose Tri-branch Neural Fusion (TNF) for multimodal data classification shown in Fig. \ref{Totalnet}. Our method takes advantage of both ensemble and fusion. A medical image classification model classifies the medical image to obtain likelihood $z_{\rm{i}}$. A medical tabular classification model classifies medical tabular to obtain likelihood $z_{\rm{t}}$. At the same time, we fuse the features from the medical image classification model and the features from the medical tabular classification model to obtain likelihood $z_{\rm{f}}$. We ensemble likelihood $z_{\rm{i}}$, $z_{\rm{t}}$ and $z_{\rm{f}}$ to get the classification result. 

Our TNF does not rely on a specific network structure. In the experiments, we validated our method by using a variety of pure convolutional neural network (CNN) structures and CNN+Transformer~\cite{vaswani2017attention} hybrid structures. We also validated our method on two image-tabular multimodal datasets: the pulmonary embolism (PE) dataset~\cite{zhou2021radfusion} which consists of PE CT scans and clinical records; as well as the cognitive impairment level classification dataset~\cite{beekly2004national} which consists of brain MRI scans and questionnaire forms. Our method achieved superior performance than the ensemble method and the fusion method, improving the ACC, MCC~\cite{chicco2021matthews}, AUROC and other metrics by 1\% to 5\%. 
We also observed that, benefited from the fine-tuning process of TNF, the image branch and the tabular branch of TNF yielded better results than their single-modal counterparts.
Additionally, during the inference stage, the TNF model can work even if one modality (either image or tabular) is missing. This shows TNF's flexibility over traditional multimodal fusion methods, which require all types of data to be present for making predictions.

The contributions of this paper can be summarized as:
\begin{itemize}
    \item We propose Tri-branch Neural Fusion (TNF), a high-performance classification structure that combines fusion and ensemble for multimodal medical data classification.
    \item We further propose two approaches named label masking and maximum likelihood selection to tackle the problem of label inconsistency in a multimodal classification task. These two methods achieved satisfying results on TNF-based models.
    \item Sufficient experiments on various Transformer and CNN structures proves that TNF is superior to individual fusion or ensemble. Experiments on multiple datasets demonstrate the generality of TNF.
\end{itemize}


\begin{figure*}[tb]
    \centering
    \includegraphics[width=0.95\textwidth]{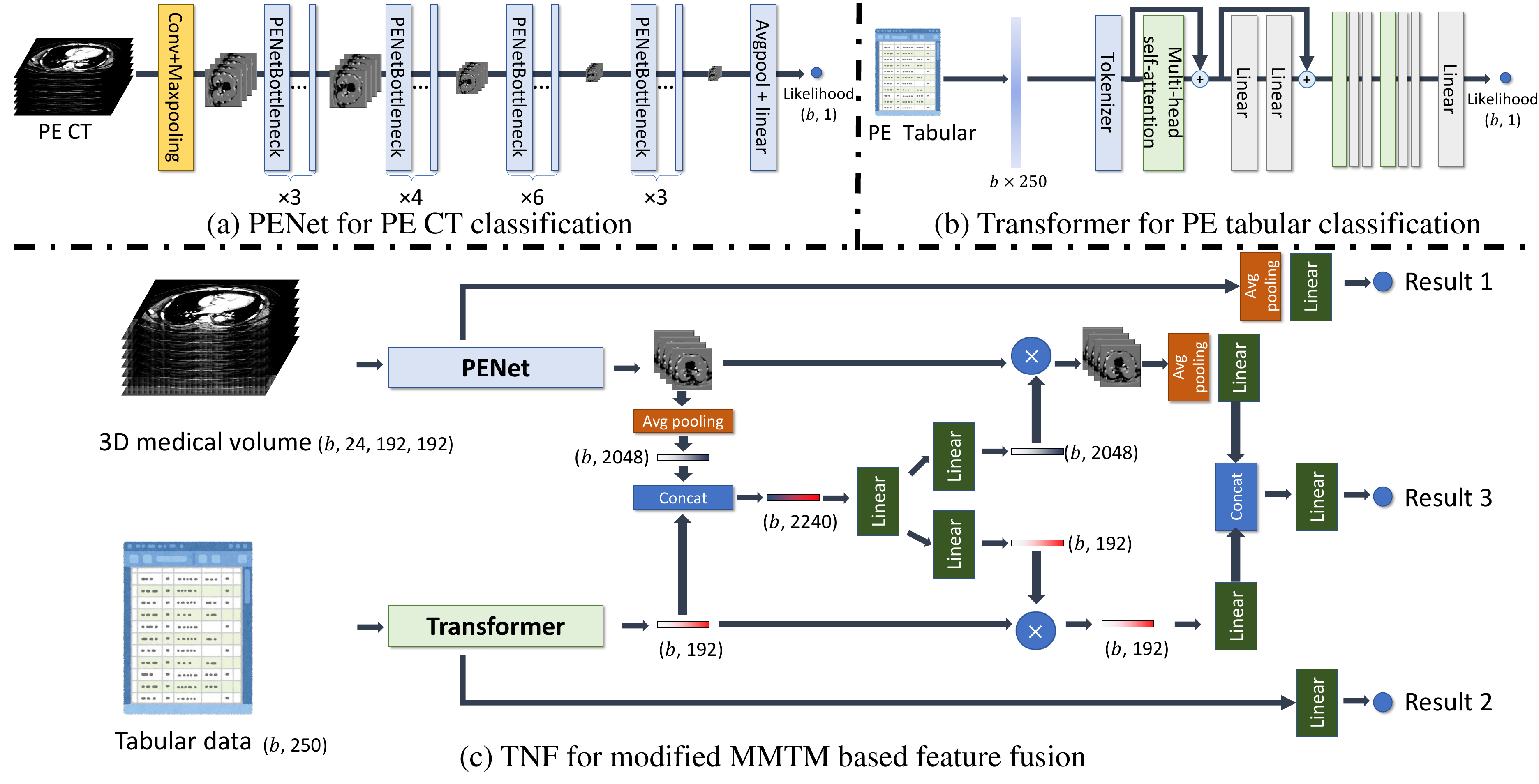}
    \caption{Classification models used in TNF for pulmonary embolism (PE) classification.}
    \label{PEFusion_network}
\end{figure*}

\section{Related Works}
\subsection{Multi-modality in Medical Data Processing}
A modality is a particular mode wherein something exists, is experienced, or is expressed~\cite{pei2023review}. For example, data regarding two datasets collected under two different circumstances, can be regarded as two modalities~\cite{baltruvsaitis2018multimodal}. 
In the field of machine learning for medical data, efforts have been made to emulate the multimodal nature of clinical expert decision-making to enhance prediction accuracy~\cite{su2020deep}.
The multi-modality medical data processing mainly focuses on classification, prediction and segmentation. 

About classification, MRI and tabular data was utilized together for breast cancer classification~\cite{holste2021end}. Notably, the data in this study was exclusively collected from a single institution. Another study employed multiple imaging modalities for skin lesion classification~\cite{yap2018multimodal}. However, its accuracy improvement over the baseline was found to be limited. About prediction, histology pathological images and genomics information were combined together for survival prediction~\cite{mobadersany2018predicting}. However, the dataset used in this study is relatively small. Additionally, Brain lesion patterns and user-defined clinical measurements were used for predicting conversion to multiple sclerosis~\cite{yoo2019deep}. This method needs user-defined MRI measurements, necessitating collaboration with clinicians. 


It is noteworthy that the aforementioned methods share three common limitations. First, they primarily concentrate on enhancing specific network structures, which may lack versatility across different machine learning architectures (e.g. unable to apply improvements designed for CNNs to Transformer models). Second, labels can vary between modalities. However, these methods require the labels of different modalities to be consistent for the training process. Third, these methods necessitate the availability of multi-modal inputs both during training and inference stages. If any modality is missing during the inference phase, these methods become inoperable. Our innovative approach effectively overcomes these three limitations.

\subsection{Ensemble and Fusion Techniques in Multimodal Classification}
Ensemble and fusion are two prevalent strategies in multimodal data classification. The ensemble approach combines outputs from multiple base classifiers, enhancing robustness and accuracy. This technique has been proven effective in RGB nature image classification~\cite{beluch2018power} and medical image classification~\cite{zhou2021radfusion}.

In contrast, fusion integrates data from multiple sources or modalities at different levels to enhance classification performance. It has been extensively applied in deep learning-based classification. In the context of CNN, the Multimodal Transfer Module (MMTM)~\cite{joze2020mmtm} implements fusion of feature modality within convolution layers in varying spatial dimensions. Weighted Feature Fusion~\cite{dong2022weighted} merges features from Graph Attention Networks and CNNs for hyperspectral image classification. Auxiliary supervision~\cite{holste2023improved} generates additional sources during training to increase fusion on small datasets. Regarding Transformer-based fusion, Cross-Modal Attention~\cite{li2020multimodal} facilitates simultaneous attention to two distinct sequences or feature sets. TokenFusion~\cite{wang2022multimodal} identifies uninformative tokens, replacing them with projected and aggregated inter-modal features in multimodal classification tasks.

\section{Methodology}
\subsection{Overview}
Given a medical image represented as $\boldsymbol{x}_{\rm{i}}$ and a set of tabular attributes denoted as $\boldsymbol{x}_{\rm{t}}$, our approach yields three distinct outputs: $z_{\rm{i}}$, $z_{\rm{t}}$, and $z_{\rm{f}}$, all derived from a deep learning model denoted as $\mathcal{F}(\cdot)$. Here, $z_{\rm{i}}$ and $z_{\rm{t}}$ is derived from the medical image and the tabular attributes, respectively, and $z_{\rm{f}}$ corresponds to the fused features of both the medical image and tabular attributes. The relationship can be described as follows:
\begin{align}
    z_{\rm{i}}, z_{\rm{t}}, z_{\rm{f}} = \mathcal{F}(\boldsymbol{x}_{\rm{i}}, \boldsymbol{x}_{\rm{t}}),
\end{align}
during the training phase, the model $\mathcal{F}(\cdot)$ is optimized using a loss function $\mathcal{L}$, defined as:

\begin{equation}
    \mathcal{L} = \lambda_1 \mathcal{L}_{\rm{c}}(z_{\rm{i}}, y_{\rm{i}}) + \lambda_2 \mathcal{L}_{\rm{c}}(z_{\rm{t}}, y_{\rm{t}}) + \lambda_3 \mathcal{L}_{\rm{c}}(z_{\rm{f}}, y_{\rm{f}}),
\end{equation}
where $\lambda_1$, $\lambda_2$, and $\lambda_3$ are the weighting coefficients, $y_{\rm{i}}$, $y_{\rm{t}}$ and $y_{\rm{f}}$ are labels of the image, tabular and fusion branch, and $\mathcal{L}_{\rm{c}}(\cdot)$ represents the cross-entropy loss. In the inference phase, the classification result is determined by:

\begin{equation}
    \hat{y} = \mathcal{G}\left(\frac{z_{\rm{i}} + z_{\rm{t}} + z_{\rm{f}}}{3}\right),
\end{equation}
with $\mathcal{G}(\cdot)$ being a threshold function defined by a threshold $\theta$. For binary classification scenarios, this function is expressed as:

\begin{align}
    \mathcal{G}(z) &= 1, \quad \text{if } \alpha \geq \theta, \\
    \mathcal{G}(z) &= 0, \quad \text{if } \alpha < \theta,
\end{align}
and commonly $\theta=0.5$. The complete framework of this method is illustrated in Fig. ~\ref{Totalnet}.

\begin{figure*}[tb]
    \centering
    \includegraphics[width=0.90\textwidth]{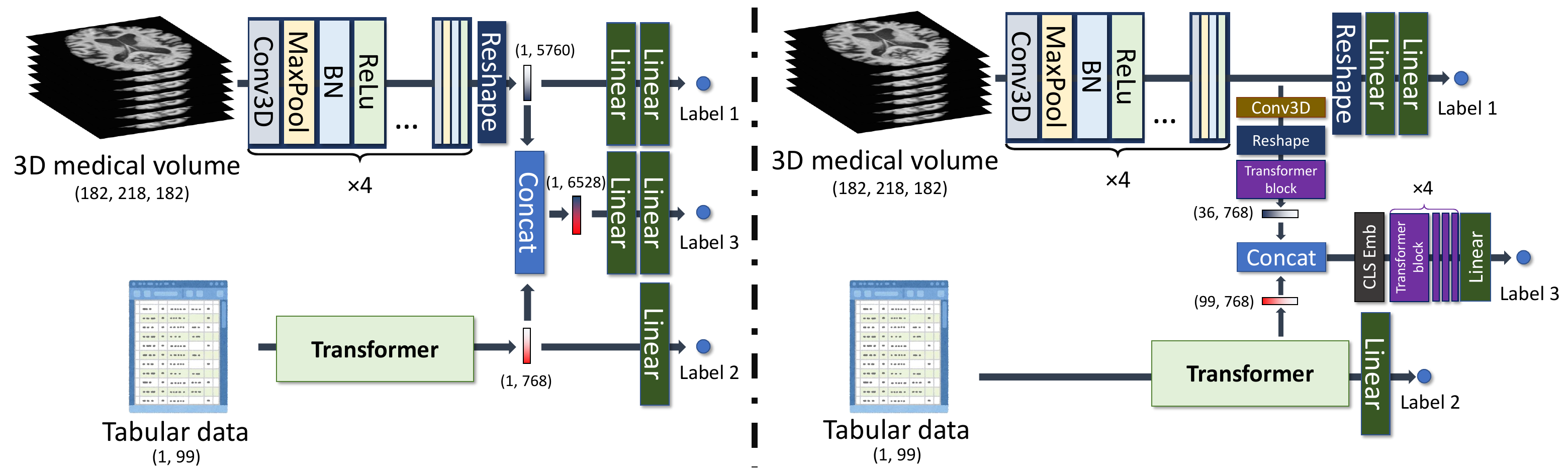}
    \caption{Models utilized in TNF for cognitive impairment level classification on the NACC dataset.}
    \label{NACCFusion}
\end{figure*}

\subsection{TNF for Pulmonary Embolism (PE) Classification}
\label{Sec_PE_method}
\subsubsection{Backbone}
We employed the Stanford AIMI pulmonary embolism (PE) dataset~\cite{zhou2021radfusion} as the first dataset in our study. This dataset comprises computed tomography (CT) scans and corresponding clinical data extracted from the electronic health records (EHR) of the same patients. We chose benchmark architectures: for the classification of CT scans, we utilized the PENet architecture~\cite{huang2020penet}, and for the EHR data classification, we employed the Tabular Transformer~\cite{NEURIPS2021_9d86d83f}. The integration of these methods in our framework is depicted in Fig. \ref{PEFusion_network}.

\subsubsection{Training with Inconsistent Labels in Multimodal Fusion}
Addressing label inconsistency is crucial in multimodal classification. In the PE dataset, each CT scan slice has its own label. For instance, a CT volume of size 256$\times$256$\times$100 yields 100 individual slice labels. However, the corresponding EHR data, is assigned only one single label, either positive or negative. 
Previous study~\cite{huang2020penet} established a benchmark for classifying PE dataset image data, dividing the CT volume into multiple 24 consecutive slices. If any 4 slices within any 24 consecutive slices are identified as positive, the entire volume is labeled positive. This approach does not completely solve the label inconsistency problem when fusing EHR and CT volume data, as the EHR label might not align with the image's. Refer to Appendix A for details.

To resolve this discrepancy, we propose two approaches: 1) label masking, and 2) maximum likelihood selection.

\noindent \textbf{$\bullet$ Label Masking} The implementation of label masking can be described as;
\begin{equation}
    \mathcal{L} = 
    \begin{cases}
        \lambda_1 \mathcal{L}_{\rm{c}}(z_{\rm{i}}, y_{\rm{i}}) + \lambda_2 \mathcal{L}_{\rm{c}}(z_{\rm{t}}, y_{\rm{t}}) + \lambda_3 \mathcal{L}_{\rm{c}}(z_{\rm{f}}, y_{\rm{i}}), & \text{if } y_{\rm{i}} = y_{\rm{t}}, \\
        \lambda_1 \mathcal{L}_{\rm{c}}(z_{\rm{i}}, y_{\rm{i}}) + \lambda_2 \mathcal{L}_{\rm{c}}(z_{\rm{t}}, y_{\rm{t}}), & \text{otherwise},
    \end{cases}
\end{equation}
where $y_{\rm{i}}$ and $y_{\rm{t}}$ are labels of image and tabular, respectively. The fusion branch's weights will be optimized only if $y_{\rm{i}}=y_{\rm{t}}$. With label masking, the training of fusion branch will not be conducted if the labels are inconsistent.  

\noindent \textbf{$\bullet$ Maximum Likelihood Selection}
We use pre-trained PENet in previous research~\cite{zhou2021radfusion} to extract the 24 slices with the highest likelihood for each volume, and use these 24 slices as input to train TNF-based models. If the entire volume is positive, then these 24 slices are also positive, otherwise they are negative. 

This process can be described as,

\begin{equation}
    \begin{aligned}
        &j' = \argmax_j(\mathcal{F}_{\rm{i}}(\boldsymbol{x}^{j}_{\rm{i}})), \quad j \in \{1, \ldots, N\}, \\
        &z_{\rm{i}}, z_{\rm{t}}, z_{\rm{f}} = \mathcal{F}(\boldsymbol{x}^{j'}_{\rm{i}}, \boldsymbol{x}_{\rm{t}}),
    \end{aligned}
\end{equation}
where \(N\) represents the total number of 24-slice groups within a CT volume. For example, if a CT volume contains 100 slices, 
$N = \left \lceil{\frac{100}{24}}\right \rceil = 5$.
This means there are five groups, with Groups 1 to 4 each comprising 24 slices, and Group 5 containing only 4 slices. We use padding by filling Group 5 with the CT value of air (-1000), expanding it from 4 to 24 slices. $\mathcal{F}_{\rm{i}}$ is the PENet in Fig.~\ref{PEFusion_network}~(a). By only selecting the slices that with the highest positive likelihood, maximum likelihood selection solves the inconsistency problem between image and tabular labels.

\subsubsection{TNF-Based Models for PE Classification}
We validated several models based on TNF for PE classification.

\noindent\textbf{$\bullet$ MMTM-based Feature Fusion}
First, we conducted experiments based on TNF for MMTM-based feature fusion. MMTM~\cite{joze2020mmtm} is a widely-used feature fusing method for fusing the knowledge of multiple modalities like

\begin{equation}
\begin{split}
    \boldsymbol{v}_{\rm{a}}' =& \boldsymbol{v}_{\rm{a}} \odot f_{\rm{a}}(f_{\rm{c}}(C(\mathcal{S}(\boldsymbol{v}_{\rm{a}}), \mathcal{S}(\boldsymbol{v}_{\rm{b}}))), \\
    \boldsymbol{v}_{\rm{b}}' =& \boldsymbol{v}_{\rm{b}} \odot f_{\rm{b}}(f_{\rm{c}}(C(\mathcal{S}(\boldsymbol{v}_{\rm{a}}), \mathcal{S}(\boldsymbol{v}_{\rm{b}}))),
\end{split}
\end{equation}
where $\mathcal{S}$ is the ``squeeze'' function to squeeze multi-dimension feature $\boldsymbol{v}_{\rm{a}}, \in \mathbb{R}^{b \times c \times m_1,...,m_K}$ to $\mathbb{R}^{b \times c}$. $f_{\rm{a}}(\cdot)$ and $ f_{\rm{b}}(\cdot)$ and $f_{\rm{c}}(\cdot)$ are linear layers respectively. $C(\cdot)$ is channel-wise concatenation. $\odot$ is channel-wise multiply.

In our context, features $\boldsymbol{v}_{\rm{i}} \in \mathbb{R}^{b \times c \times h \times w \times d}$ are obtained from a 3D medical volume, while the features $\boldsymbol{v}_{\rm{t}} \in \mathbb{R}^{b \times l}$ are derived from a Tabular Transformer. Due to the differing dimensions of these features, where $\boldsymbol{v}_{\rm{t}}$ from the Tabular Transformer has only two dimensions, squeezing $\boldsymbol{v}_{\rm{t}}$ is deemed unnecessary. Consequently, we adapted the MMTM method to fuse $\boldsymbol{v}_{\rm{i}}$ and $\boldsymbol{v}_{\rm{t}}$, as illustrated in Fig. \ref{PEFusion_network} (c). The modified fusion process is defined as:

\begin{equation}
\begin{split}
    \boldsymbol{v}_{\rm{i}}' =& \boldsymbol{v}_{\rm{i}} \odot f_{\rm{i}}(f_{\rm{c}}(C(f_{\rm{avg}}(\boldsymbol{v}_{\rm{i}}), \boldsymbol{v}_{\rm{t}}))), \\
    \boldsymbol{v}_{\rm{t}}' =& \boldsymbol{v}_{\rm{t}} \odot f_{\rm{t}}(f_{\rm{c}}(C(f_{\rm{avg}}(\boldsymbol{v}_{\rm{i}}), \boldsymbol{v}_{\rm{t}}))),
\end{split}
\end{equation}
where $f_{\rm{avg}}$ is average pooling function. Subsequently, the fused output is obtained through a process of average pooling, concatenation, and application of a linear layer, as depicted in Fig. \ref{PEFusion_network} (c).

\noindent \textbf{$\bullet$ Concatenation-Based Transformer Feature Fusion}
The effectiveness of TNF was also evaluated on Transformer structures. For image features $\boldsymbol{v}_{\rm{i}} \in \mathbb{R}^{b \times c \times h \times w \times d}$, we apply 3D convolution and a reshaping function to transform them into $\boldsymbol{v}_{\rm{i}}' \in \mathbb{R}^{b \times h \times w \times d \times c'}$. These features are then concatenated channel-wise with outputs from the Tabular Transformer. The likelihood is calculated as follows:

\begin{equation}
\begin{split}
    z_f = \mathcal{T}(C(\boldsymbol{v}_{\rm{i}}', \boldsymbol{v}_{\rm{t}})),
\end{split}
\end{equation}

where $\mathcal{T}$ represents the class embedding and Transformer blocks. Refer to Appendix B for detailed network structure.

\noindent \textbf{$\bullet$ TokenFusion-Based Transformer Feature Fusion:}
TNF also demonstrates effectiveness with TokenFusion~\cite{wang2022multimodal} and Cross-Modal Attention~\cite{li2020multimodal} in Transformer feature fusion. We use TokenFusion to fuse features from image and tabular modalities:

\begin{equation}
\begin{aligned}    
    \boldsymbol{v}_{\rm{i}}' &= \boldsymbol{v}_{\rm{i}} + \rm{MLP}((\rm{SA}(\boldsymbol{v}_{\rm{i}}) \odot \it{f}(\boldsymbol{v}_{\rm{i}})) + \boldsymbol{v}_{\rm{i}}), \\
    \boldsymbol{v}_{\rm{t}}' &= \boldsymbol{v}_{\rm{t}} + \rm{MLP}((\rm{SA}(\boldsymbol{v}_{\rm{t}}) \odot \it{f}(\boldsymbol{v}_{\rm{t}})) + \boldsymbol{v}_{\rm{t}}),
\end{aligned}
\end{equation}
where $\rm{MLP}$ is a multi-layer perceptron, $\rm{SA}$ is a self-attention mechanism, and $f(\cdot)$ is a linear function. Afterwards, $\boldsymbol{v}_{\rm{i}}'$ and $\boldsymbol{v}_{\rm{t}}'$ will be fused to get the classification result. TokenFusion assigns weights to each token, selectively emphasizing or de-emphasizing certain tokens.

\noindent \textbf{$\bullet$ Cross-Modal Attention-Based Transformer Feature Fusion:}
On the other hand, Cross-Modal Attention facilitates simultaneous attention to two distinct feature sources:

\begin{equation}
\begin{aligned}
    \boldsymbol{v}_{\rm{i}}' &= \rm{MA}(\mathcal{Q}_{t}(\boldsymbol{v}_{\rm{t}}), \mathcal{K}_{i}(\boldsymbol{v}_{\rm{i}}), \mathcal{V}_{i}(\boldsymbol{v}_{\rm{i}})),
    \\
    \boldsymbol{v}_{\rm{t}}' &= \rm{MA}(\mathcal{Q}_{i}(\boldsymbol{v}_{\rm{i}}), \mathcal{K}_{t}(\boldsymbol{v}_{\rm{t}}), \mathcal{V}_{t}(\boldsymbol{v}_{\rm{t}})),
\end{aligned}
\end{equation}
where $\rm{MA}$ represents a multi-head attention mechanism, while $\mathcal{Q}$, $\mathcal{K}$, and $\mathcal{V}$ denote the query, key, and value operations.

\subsection{TNF for Cognitive Impairment Level Classification}
\label{Sec_NACC_method}
TNF was also employed for a multi-class classification task involving cognitive impairment level classification using the NACC~\cite{beekly2004national} dataset. The cognitive impairment levels are categorized into three stages: 0 (normal), 1 (mild cognitive impairment), and 2 (severe cognitive impairment). The NACC dataset comprises two types of modalities: brain MRI volumes (image data) and questionnaire forms (tabular data). We developed two models based on TNF for classifying NACC data.

\subsubsection{Concatenation-Based Feature Fusion}
The first model, depicted in Fig.~\ref{NACCFusion}~(a), employs a CNN for classifying brain MRI volumes and a Tabular Transformer for processing tabular data. We concatenate the features from the brain MRI volume, with a shape of $(b, 5760)$, with those from the Tabular Transformer, with a shape of $(b, 768)$. This results in a combined feature vector of shape $(b, 6528)$, which is then fed into linear layers to derive the fusion label.

\subsubsection{Transformer Feature Fusion}
The second model, shown in Fig.~\ref{NACCFusion}~(b), involves processing features from the brain MRI volume through a 3D convolution layer, a reshaping operation, and a Transformer block. This process yields a feature vector of shape $(b, 36, 768)$. Features from the intermediate layers of the Tabular Transformer, shaped $(b, 99, 768)$, are then concatenated with this vector to form a new vector of shape $(b, 135, 768)$. This combined vector is subsequently inputted into Transformer layers to obtain the fusion label.



\begin{table}[t]
\resizebox{0.49\textwidth}{!}{
\begin{tabular}{c|cc|cc}
\hline
       &  \multicolumn{2}{c|}{Label masking} & \multicolumn{2}{c}{Maximum likelihood selection} \\ \hline
       & w/o Pre-train            & w/ pre-train            & w/o Pre-train   & w/ pre-train  \\ \hline
AUPRC $\uparrow$ & 0.871 & 0.887 & 0.844 & \textbf{0.910} \\
AUROC $\uparrow$ & 0.818 & 0.845 & 0.797 & \textbf{0.867} \\
ACC $\uparrow$   & 0.747 & 0.768 & 0.742 & \textbf{0.789} \\
MCC $\uparrow$   & 0.477 & 0.529 & 0.464 & \textbf{0.582} \\
Recall $\uparrow$ & \textbf{0.891} & 0.782 & 0.827 & 0.764 \\ \hline
\end{tabular}}
\caption{Comparison of maximum likelihood selection and label masking on PE dataset w/ and w/o pre-training.}
\label{Comparison_likelihoodmask}
\end{table}

\begin{table*}[t]
\centering
\resizebox{\textwidth}{!}{\begin{threeparttable}\begin{tabular}{c|cc|cc|c|cccc}
\hline
       & \multicolumn{2}{c|}{Single modal} & \multicolumn{2}{c|}{Multimodal} & Ensemble       & \multicolumn{4}{c}{TNF + maximum likelihood selection}                           \\ \hline
       & Img only           & Tab only     & MMTM fusion$^1$       & Clip fusion       & Img+Tab & MMTM  & Concatenation &  TokenFusion & CM Attention$^2$  \\ \hline
AUPRC $\uparrow$  & 0.848              & 0.842        & 0.861            & 0.838               & 0.888          & \textbf{0.910} & 0.907        & 0.905         & 0.895          \\
AUROC $\uparrow$ & 0.786              & 0.788        & 0.816            & 0.797               & 0.847          & \textbf{0.867} & 0.865        & 0.860         & 0.848          \\
ACC $\uparrow$   & 0.716              & 0.737        & 0.721            & 0.705               & 0.774          & \textbf{0.789} & 0.779        & 0.784         & \textbf{0.789} \\
MCC $\uparrow$   & 0.407              & 0.472        & 0.455            & 0.393               & 0.534          & \textbf{0.582} & 0.567        & 0.567         & 0.567          \\
Recall $\uparrow$ & 0.855     & 0.727        & 0.673            & \textbf{0.918}               & 0.818          & 0.764          & 0.736        & 0.773         & 0.827   \\   \hline
\end{tabular}
\begin{tablenotes}
    \footnotesize
    \item[1] Structure is equivalent to the model in Fig.~\ref{PEFusion_network}~(c) while only outputs fusion result.
    \item[2] ``Cross-Modal Attention-Based Transformer Feature Fusion" in Section~\ref{Sec_PE_method}.
\end{tablenotes}
\end{threeparttable}}
\caption{Results on PE dataset.}
\label{Tab_PEResults}
\end{table*}

\begin{table}[t]
\resizebox{0.49\textwidth}{!}{
\begin{threeparttable}
\begin{tabular}{c|cc|c|c|cc}
\hline
        & \multicolumn{2}{c|}{Single modal} & Multimodal & Ensemble & \multicolumn{2}{c}{TNF} \\ \hline
        & Img             & Tab            & Fusion & Img+Tab & Model 1$^1$             & Model 2$^2$            \\ \hline
ACC $\uparrow$    & 0.669           & 0.761          & 0.653 & 0.785            & 0.785               & \textbf{0.805}              \\
MCC $\uparrow$    & 0.509           & 0.564          & 0.381 & 0.623           & 0.629               & \textbf{0.672}              \\
Recall $\uparrow$ & 0.657           & 0.624         & 0.488 & 0.667           & 0.682               & \textbf{0.712}              \\
Jaccard $\uparrow$ & 0.480           & 0.516          & 0.342 & 0.544           & 0.552               & \textbf{0.580}              \\
F1 $\uparrow$     & 0.640           & 0.655          & 0.431 & 0.677          & 0.686               & \textbf{0.709}      \\
\hline
\end{tabular}
\begin{tablenotes}
    \footnotesize
    \item[12] ``Concatenation-Based Feature Fusion" and ``Transformer Feature Fusion" models in Section~\ref{Sec_NACC_method}, respectively.
\end{tablenotes}
\end{threeparttable}}
\caption{Results on NACC dataset. Since AUPRC and AUROC are not applicable to a multi-classification problem, we added Jaccard and macro F1 score as additional metrics.}
\label{Tab_NACCResults}
\end{table}


\begin{table}[t]
\resizebox{0.49\textwidth}{!}{
\begin{threeparttable}\begin{tabular}{c|ccccc}
\hline
       & Img branch & Tab branch & Fusion & Ensemble$^1$ & TNF        \\ \hline
AUPRC $\uparrow$ & 0.864       & 0.840    & 0.849    & \textbf{0.910}        & \textbf{0.910}           \\
AUROC $\uparrow$ & 0.792       & 0.790    & 0.799    & 0.865        & \textbf{0.867}           \\
ACC $\uparrow$   & 0.684       & 0.705    & 0.768    & 0.768        & \textbf{0.789}           \\
MCC $\uparrow$   & 0.437       & 0.405    & 0.474    & 0.561        & \textbf{0.582}           \\
Recall $\uparrow$ & 0.536      & 0.709    & \textbf{0.764}    & 0.691        & \textbf{0.764}           \\ \hline
\end{tabular}
\begin{tablenotes}
    \footnotesize
    \item[1] The ensemble of Img and Tab branch after TNF fine-tuning.
\end{tablenotes}
\end{threeparttable}}
\caption{Performance of each branch of TNF models on PE dataset.}
\label{Tab_PEAblation}
\end{table}
\section{Experiments and Results}

\subsection{Preprocessing}
\subsubsection{Preprocessing of PE Dataset}
The tabular of PE dataset contains 1,505 attributes such as age, gender, pulse, etc. We remove two attributes that are directly related to PE (``DISEASES OF PULMONARY CIRCULATION: frequency'' and ``DISEASES OF PULMONARY CIRCULATION: presence'') because we found if these two attributes are not 0, PE must be positive. We use SHAP~\cite{shrikumar2017learning} to select the most important 250 attributes for PE classification.

We also resized images from PE dataset from size 256$\times$256 to 224$\times$224, and center-cropped their size to 192$\times$192. We normalized the image intensity to range [-1,~1].

\subsubsection{Preprocessing of NACC Dataset}

\noindent \textbf{$\bullet$ Case Selection and Image Preprocessing}
The NACC dataset contains 9,812 case of brain MRI with T1, T2 and FLAIR modality. We select thin-slice cases that have T1 modality and use FreeSurfer~\cite{fischl2012freesurfer} to perform bone removal, resize and intensity normalization. After preprocessing, there remain 1,252 cases with size 182$\times$218$\times$182 voxels. We normalized the volume intensity to range [-8,~8].

\noindent \textbf{$\bullet$ Tabular Preprocessing} 
We preprocessed NACC tabular data by zero padding, feature selection and normalization. After preprocessing, the tabular data was left with 99 features including cognitive questionnaire information, gender, age, etc.

\subsection{Training and Inference}
\subsubsection{PE Dataset}
We follow the experimental settings of the PE dataset~\cite{zhou2021radfusion}. There are totally 1,837 cases, 1,454 cases for training, 190 cases for validation, and 193 cases for testing.

We loaded the best pre-trained PENet provided by prior research~\cite{huang2020penet}. To ascertain the saturation of the pre-trained weights, we ran additional training epochs starting from the pre-trained weights, and confirmed no further improvement in performance; we loaded pre-trained Tabular Transformer trained by ourselves, to TNF-based models for training.
The learning rate is 10$^{-4}$ with cosine decay down to 10$^{-5}$. The number of epoch is 1,000, the optimizer is AdamW~\cite{loshchilov2018decoupled}. The regularization parameters are set as $\lambda_1 = 0.1$, $\lambda_2 = 0.1$, $\lambda_3 = 0.8$. The best model on the validation dataset was used for testing.
 
\subsubsection{NACC Dataset}
Among the 1,252 cases of the NACC dataset, we randomly selected 752 cases for training, 249 cases for validation and 251 cases for testing.

We pre-trained a 4-block CNN model consistent with the architecture used in previous research~\cite{qiu2022multimodal}, as shown in Fig.~\ref{NACCFusion}, for image classification; and a Tabular Transformer model with the same structure in Fig.~\ref{PEFusion_network}~(b) for tabular classification. After pre-training, we train TNF-based models shown in  Fig.~\ref{NACCFusion}~(a)~(b). The learning rate is 10$^{-3}$, epoch is 100, and the optimizer is Adam. 

\subsection{Results}
\subsubsection{PE Dataset}
First, we evaluated two approaches on the PE dataset: training with inconsistent labels using maximum likelihood selection against label masking, and comparing training from scratch against pre-training. For the case of training from scratch, only the PENet was pre-trained on the Kinetics-600 dataset~\cite{carreira2018short}. For the case of pre-training, both the PENet and the Tabular Transformer were pre-trained on the PE dataset. The outcomes of these comparisons are presented in Table~\ref{Comparison_likelihoodmask}. We found that the combination of maximum likelihood selection and pre-training yielded the best performance across most metrics. Therefore, we select it for the following experiments.

\begin{figure}[t]
  \centering
  \includegraphics[width=0.49\textwidth]{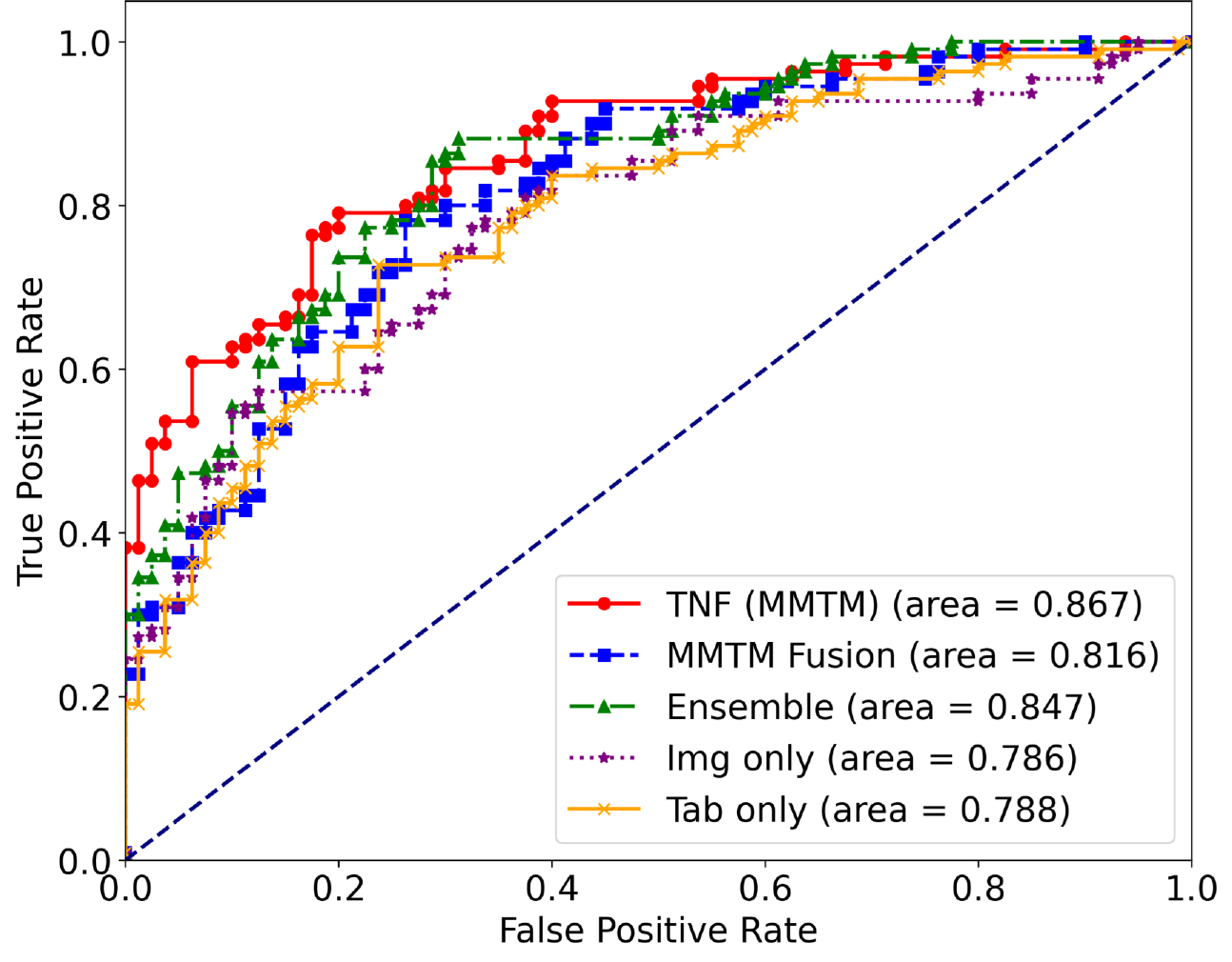}
  \caption{ROC curve of TNF and other methods.}
  \label{Roccurve}
\end{figure}

The results for the PE dataset are detailed in Table~\ref{Tab_PEResults}, with the ROC curve displayed in Fig.~\ref{Roccurve}. We compared TNF-based models with single modal, multimodal fusion, and ensemble approaches. We tried two fusion models: 1) MMTM fusion, whose structure is equivalent to the model in Fig.~\ref{PEFusion_network}~(c) while only outputs fusion result. 2) Clip fusion, based on the research of~\cite{Hager_2023_CVPR} (refer to Appendix C for details). The TNF approach not only surpassed single-modal classification (using either image or tabular) but also outperformed multimodal fusion and ensemble in most metrics. Notably, the fusion approach demonstrated lower accuracy compared to the ensemble method, indicating that conventional multimodal fusion may be ineffective for some datasets.

\subsubsection{NACC Dataset}
The quantitative results for the NACC dataset are presented in Table~\ref{Tab_NACCResults}. For comparison, the multimodal fusion model is identical to the model with only the fusion branch, as shown on the left in Fig.~\ref{NACCFusion}. Our two TNF-based models, namely the simple concatenation-based feature fusion and the Transformer feature fusion model, demonstrated superior performance compared to other methods.

\begin{figure}[t]
  \centering
  \includegraphics[width=0.49\textwidth]{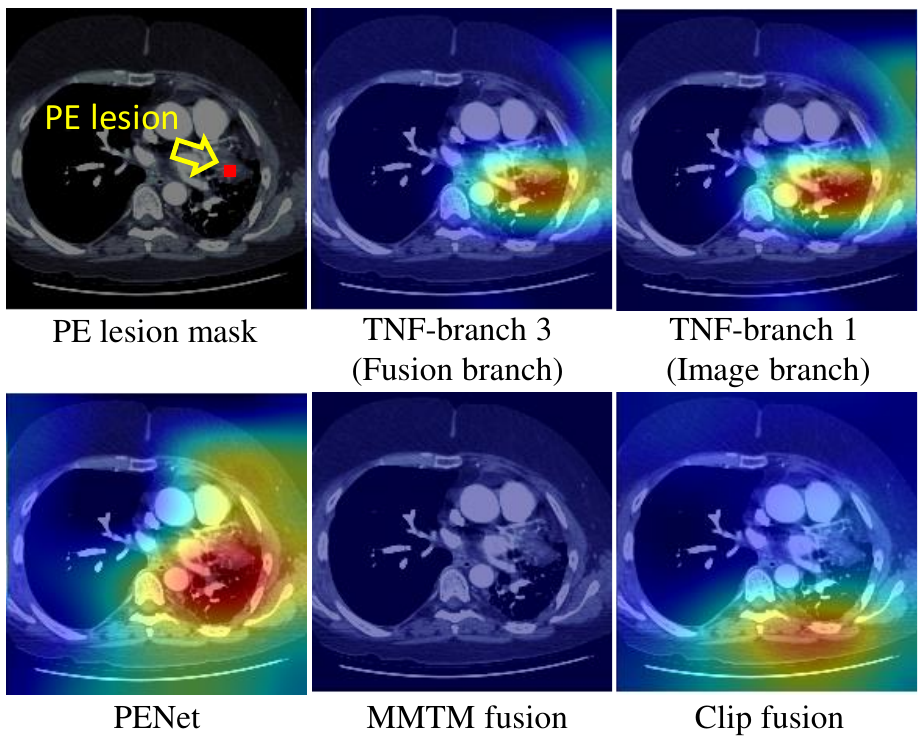}
  \caption{Grad-CAM heatmaps (zoom in for better observation).}
  \label{GradCAM}
\end{figure}

\subsection{3D Grad-CAM-based Critical Region Analysis}

To visualize the critical regions for the models' predictions, we designed a 3D Grad-CAM model and adapted it to the TNF models. Grad-CAM~\cite{selvaraju2017grad} produces visual explanations for decisions made by CNNs. We modified Grad-CAM so that it can be applied to multiple inputs and multiple outputs, as well as can process 3D medical images. We back-propagate one output at a time to get its Grad-CAM heatmaps.

We compared five types of Grad-CAM heatmaps: 1) The fusion branch of TNF. the model is shown in Fig.~\ref{PEFusion_network} (c); 2) The image branch of TNF; 3) PENet; 4) MMTM fusion model; The model's structure is equivalent to the model in Fig.~\ref{PEFusion_network} (c) and only outputs fusion result; 5) Clip fusion model. 

The Grad-CAM heatmaps are compared with PE lesion masks marked by clinicians shown in Fig.~\ref{GradCAM} as an exmaple. TNF's Grad-CAM heatmap highlighted the lesion area. On the other hand, PENet's heatmap highlighted a larger area. MMTM fusion and Clip fusion failed to highlight the lesion. Generally, TNF's heatmap is the closest to the lesion mask.

\subsection{Performance of Each Branch of TNF Models}
We also conducted performance comparison of individual branches of TNF models, shown in Table~\ref{Tab_PEAblation}. Note that our TNF model has three branches: 1) the image branch; 2) the tabular branch; and 3) the fusion branch. We illustrated the results of each individual branch, the ensemble of the first two branches, as well as TNF. TNF achieved the best quantitative results, except for recall.


\section{Discussion}
\begin{table}[t]
\resizebox{0.49\textwidth}{!}{\begin{tabular}{c|cc|cc|cc}
\hline
                      & \multicolumn{2}{c|}{Tabular Transformer}                        & \multicolumn{2}{c|}{ElasticNet}                         & \multicolumn{2}{c}{ResNet}                    \\ \hline
\multicolumn{1}{c|}{} & \multicolumn{1}{c}{Tab} & \multicolumn{1}{c|}{Ensemble} & \multicolumn{1}{c}{Tab} & \multicolumn{1}{c|}{Ensemble} & \multicolumn{1}{c}{Tab} & \multicolumn{1}{c}{Ensemble} \\ \hline
AUROC $\uparrow$                & 0.788                   & 0.847                         & 0.783                   & 0.829                         & 0.750                   & 0.847                        \\
ACC $\uparrow$                  & 0.737                   & 0.774                         & 0.626                   & 0.784                         & 0.684                   & 0.716                        \\
MCC $\uparrow$                  & 0.472                   & 0.534                         & 0.349                   & 0.562                         & 0.403                   & 0.459                        \\ \hline
\end{tabular}}
\caption{Different model's performance on tabular data classification on PE dataset. Generally, Tabular Transformer is the best.}
\label{Tab_Tabnets}
\end{table}

\subsection{Comparison of Label Masking and Maximum Likelihood Selection}
To address the challenge of inconsistent labels in multimodal processing, we introduce two approaches: label masking and maximum likelihood selection. According to the results in Table \ref{Comparison_likelihoodmask}, maximum likelihood selection achieves higher AUPRC compared to label masking (0.910 vs. 0.887) when the network is pre-trained. However, without pre-training, the situation is reversed (0.844 vs. 0.871). This indicates that maximum likelihood selection is more effective for fine-tuning a pre-trained network as it does not add new label information, which might be redundant at the stage of training. Conversely, label masking is preferable for training from scratch, as it introduces a larger amount of training data. 

\subsection{Model Selection for Tabular Classification}
ElasticNet~\cite{zou2005regularization} was previously used for tabular data classification in the PE dataset. We additionally explored ResNet~\cite{jian2016deep} and the Tabular Transformer~\cite{NEURIPS2021_9d86d83f} to evaluate their effectiveness. Table \ref{Tab_Tabnets}, indicated that the Tabular Transformer outperformed other models. Consequently, we chose the Tabular Transformer as our baseline model.

\subsection{TNF Enhances Single Modal Classification}
The ``Img branch'' column in Table \ref{Tab_PEAblation} and the ``Img only'' column in Table \ref{Tab_PEResults} display the results of image classification. Following fine-tuning with the TNF, there was an observable improvement in the performance metrics for image classification. AUPRC increased from 0.848 to 0.864, and the MCC increased from 0.407 to 0.437. We attribute this enhancement to the integration of table modality features in the fusion process. TNF enables the PENet to more effectively extract common features crucial for accurate classification.

\subsection{TNF's Advantages over Multimodal Fusion}
When information from different modalities conflicts, using the fusion method will actually reduce the accuracy. As shown in Table \ref{Tab_PEResults} and Table \ref{Tab_NACCResults}, multimodal fusion cannot outperform ensemble. This is partly because if the fusion part is optimized, the feature extraction part may be poorly optimized. TNF solved this problem by extracting common features between modalities while maintaining optimization of the feature extraction part. Another advantage of TNF over fusion is that it can still infer the result while one modality is missing. For instance, if tabular data is missing, we can still infer the result by using the image branch. On the other hand, multimodal fusion models cannot infer the result while one modality's data is missing.

\section{Conclusion}
We proposed Tri-branch Neural Fusion (TNF) model for multimodal data classification. We also proposed label masking and maximum likelihood selection to address the challenge of inconsistent labels. We validated TNF model on various CNN and Transformer-based fusion algorithms on two multimodal medical datasets, all of which outperformed multimodal ensemble and fusion results. Compared with multimodal fusion, TNF can still infer the classification result when a certain mode is missing. Relative to ensemble, TNF achieved better quantitative results. Grad-CAM-based interpretability studies further proved that TNF can allow the model to focus on lesions, thus demonstrating its potential in auxiliary diagnosis. In the future, we plan to explore the case of more than two modalities fusion, as well as introduce TNF-based models into clinical practice to provide support to clinicians.



\newpage
\newpage

\setcounter{section}{0}
\renewcommand*{\theHsection}{chX.\the\value{section}}
\renewcommand\thesection{\Alph{section}}
\setcounter{figure}{0}
\renewcommand*{\theHfigure}{chX.\the\value{figure}}
\renewcommand\thefigure{\Alph{figure}}
\setcounter{table}{0}
\renewcommand*{\theHtable}{chX.\the\value{table}}
\renewcommand\thetable{\Alph{table}}
\part*{
\twocolumn[
\center{
Appendix for TNF: Tri-branch Neural Fusion for Multimodal Medical Data Classification \\
$\quad$
}
]
}

\section{Details About Maximum Likelihood Selection and Label Masking}

\subsection{Label Inconsistency Problem}
Inconsistencies may arise between EHR labels and image labels within the PE dataset as shown in Table~\ref{label_inconsis}. Huang et al.~\cite{huang2020penet} set a classification benchmark for PE image data by segmenting the CT volume into groups, with each containing 24 consecutive slices. A volume is deemed positive if at least 4 slices in any of these groups are positive. Our study adheres to this approach.

However, a positive PE volume doesn't guarantee that all its slices are positive. Actually, according to Table~\ref{label_inconsis}, for all the volumes which are annoted as positive, only 15.27\% slices are positive, and only 19.06\% 24-slice groups are positive. This discrepancy leads to label inconsistencies when comparing positive EHR data with potentially negative image groups. We propose 1) label masking, and 2) maximum likelihood selection to tackle this problem.
\subsection{Maximum Likelihood Selection}

We use pre-trained PENet in previous research~\cite{zhou2021radfusion} to extract the 24 slices with the highest likelihood for each volume, and use these 24 slices as input to train TNF-based models. The procedure is illustrared in Algorithm~\ref{Algorithm_MLS}.

\begin{algorithm}
\caption{Maximum Likelihood Selection}
\begin{algorithmic}[!htb]
\State $\mathbb{X}$: All volumes in the PE dataset
\State $\mathcal{F}$: Pre-trained PENet
\State $\boldsymbol{X}_j$: $j_{\rm{th}}$ volume of $192 \times 192 \times K$ voxels
\Procedure{MLS}{$\mathbb{X}, \mathcal{F}$}
\State $\mathbb{V} \gets [ ]$
\For{each volume $\boldsymbol{X}_{j} \in \mathbb{X}$}
    \State $L \gets [ ]$ \Comment{likelihood list}
    \State Divide $\boldsymbol{X}_j$ into $\boldsymbol{x}_{j1}, \boldsymbol{x}_{j2}, \ldots, \boldsymbol{x}_{jN}$ of size $192 \times 192 \times 24$
    \For{each $\boldsymbol{x}_{jn}$}
        \State $l \gets \mathcal{F}(\boldsymbol{x}_{jn})$ \Comment{get likelihood from PENet}
        \State Append $l$ to $L$
    \EndFor
    \State $\text{Idx} \gets \text{argmax}(L)$
    \State Append $\boldsymbol{x}_{j\text{Idx}}$ to $\mathbb{V}$
\EndFor
\State Use $\mathbb{V}$ to train TNF-based models
\EndProcedure
\end{algorithmic}
\label{Algorithm_MLS}
\end{algorithm}

\begin{algorithm}
\caption{Label Masking}
\begin{algorithmic}[!htb]
\State $\boldsymbol{x}_\text{i}, \boldsymbol{x}_\text{t}$: Image and tabular data
\State $y_\text{i}, y_\text{t}$: The corresponding labels of the image and tabular data
\State $\mathcal{L}_{\rm{c}}$: Cross-entropy loss function
\State $N$: Size of mini-batch
\Procedure{LM}{$\boldsymbol{x}_\text{i}, \boldsymbol{x}_\text{t}, y_{\text{i}}, y_{\text{t}}$}
    \For{$n = 1$ to $N$}
        \If{$y_{\text{i}}^{n} = y_{\text{t}}^{n}$}
            \State $\mathcal{L}^{n} = \lambda_1 \mathcal{L}_{\rm{c}}(z_{\rm{i}}^{n}, y_{\rm{i}}^{n}) + \lambda_2 \mathcal{L}_{\rm{c}}(z_{\rm{t}}^{n}, y_{\rm{t}}^{n}) + \lambda_3 \mathcal{L}_{\rm{c}}(z_{\rm{f}}^{n}, y_{\rm{i}}^{n})$
        \Else
            \State $\mathcal{L}^{n} = \lambda_1 \mathcal{L}_{\rm{c}}(z_{\rm{i}}^{n}, y_{\rm{i}}^{n}) + \lambda_2 \mathcal{L}_{\rm{c}}(z_{\rm{t}}^{n}, y_{\rm{t}}^{n})$
        \EndIf
    \EndFor
    \State $\mathcal{L}_{\text{total}} = \frac{1}{N} \sum_{n=1}^{N} \mathcal{L}^{n}$ \State Use $\mathcal{L}_{\text{total}}$ to train TNF-based models
\EndProcedure
\end{algorithmic}
\label{Algorithm_LM}
\end{algorithm}

\subsection{Label Masking}
With label masking, the training of fusion branch will not be conducted if the label is inconsistent. The procedure is illustrated in Algorithm 2.

For example, in a multimodal classification problem, if the labels of image and tabular data are inconsistent, it can be divided into the following two situations:

1. The labels of image and tabular are the same. Train the image, tabular, and fusion branches in the TNF-based models.

2. The labels of image and tabular are different. Only train the image and tabular branches in the TNF-based models.

\begin{table}[tb]
\centering
\resizebox{0.39\textwidth}{!}{\begin{tabular}{c|cc|cc}
\hline
      & \multicolumn{2}{c|}{Single slice} & \multicolumn{2}{c}{24 slices} \\ \hline
      & Img=1           & Img=0           & Img=1         & Img=0        \\ \hline
Tab=1 & 44,071           & 244,591          & 2,361          & 10,025        \\
Tab=0 & 0               & 433,956          & 0             & 18,608       \\ \hline
\end{tabular}}
\caption{Label inconsistency problem in the PE dataset: Each slice (image) within CT volumes is individually labeled, whereas the corresponding EHR data (tabular) for the entire CT volume has a single label, leading to situations where some image labels are 0 while the corresponding EHR data label is 1.}
\label{label_inconsis}
\end{table}

\section{Details of TNF-Based Models for PE Classification}

\begin{figure*}[tb]
  \centering
  \includegraphics[angle=90,origin=c,width=0.70\textwidth]{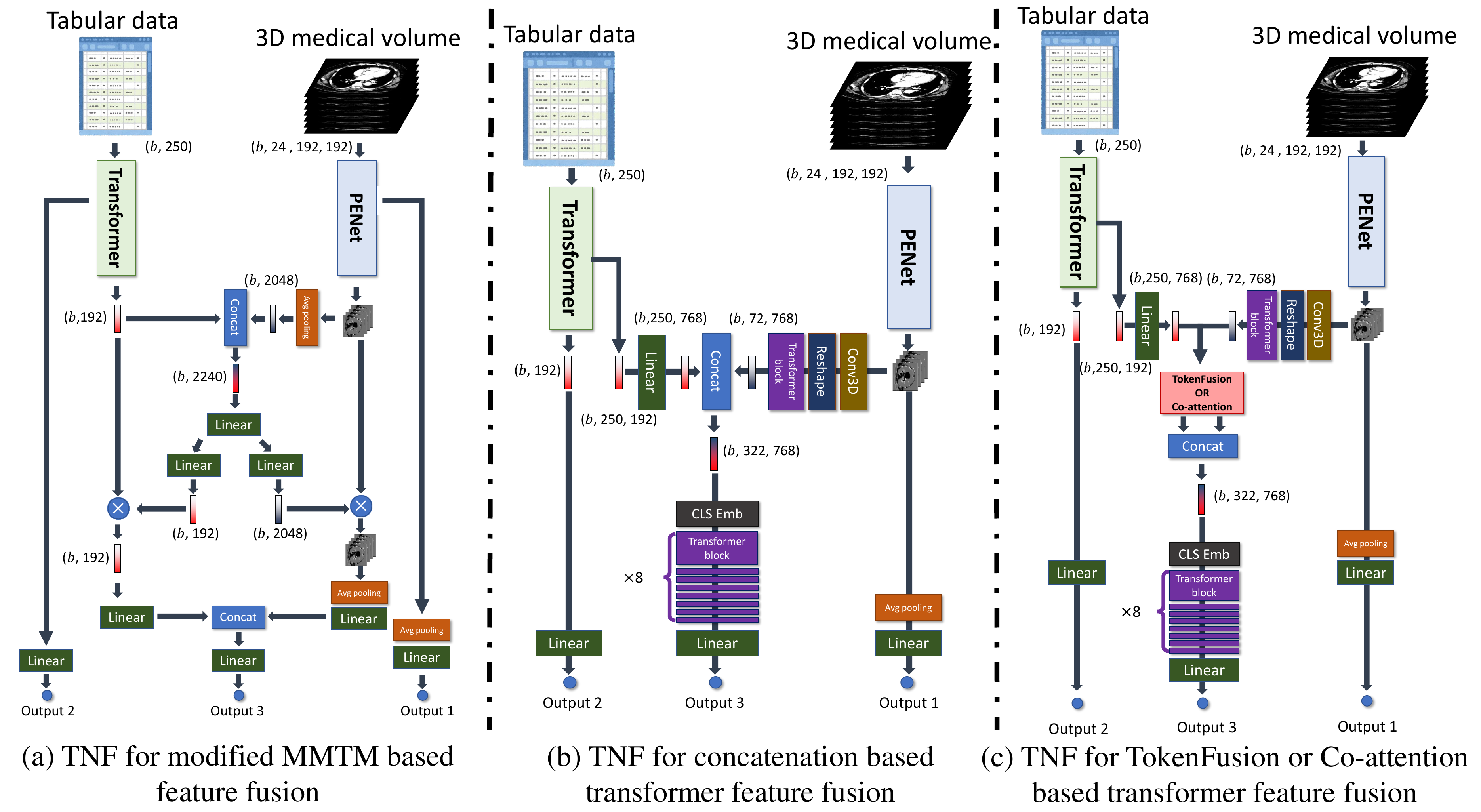}
  \caption{TNF-based networks on PE dataset.}
  \label{PE_Nets}
\end{figure*}

In order to verify the generalization of TNF, We tried serval models based on TNF for PE classification. Here we present the details of these models.

\subsection{Modified MMTM-based feature fusion}
Initially, we carried out experiments utilizing TNF with a modified MMTM-based feature fusion approach. We employed MMTM to integrate features from image and tabular data, followed by the application of a linear layer to obtain the output of the fusion branch. The architecture of this model is illustrated in Fig.~\ref{PE_Nets} (a).

\subsection{Transformer Feature Fusion}
\noindent \textbf{Concatenation-Based Feature Fusion}
We also assessed the concatenation-based feature fusion of TNF on transformer architectures. In this approach, we processed the image CNN features from PENet through a transformer block to obtain transformer features, which were then concatenated with features from the tabular transformer. Subsequently, these combined features were processed through 8 transformer blocks to derive the output of the fusion branch, as illustrated in Fig.~\ref{PE_Nets} (b).

\noindent \textbf{TokenFusion and Cross-Modal Attention-Based Feature Fusion}
The architectures of models based on TokenFusion~\cite{wang2022multimodal} and Cross-Modal Attention~\cite{li2020multimodal} are presented in Fig.~\ref{PE_Nets} (c). For fusing image and tabular features, we employed either TokenFusion or Cross-Modal Attention. These fused features were then processed through 8 transformer blocks to produce the output of the fusion branch.
\section{Details of CLIP Fusion in the ``Experiments'' Section}

\begin{figure}[!thb]
  \centering
  \includegraphics[width=0.49\textwidth]{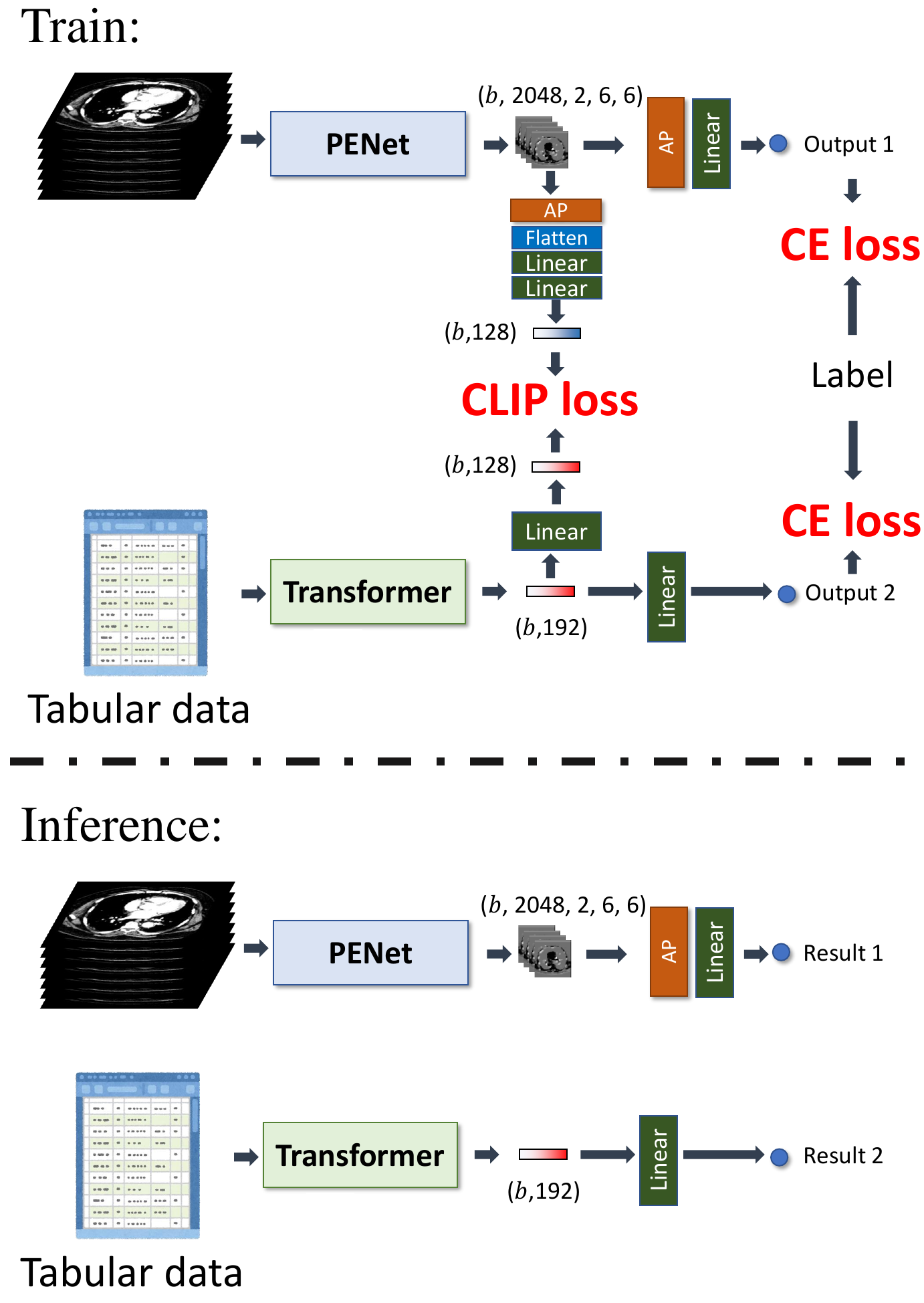}
  \caption{Process of CLIP fusion is adapted from Hager et al. \protect\cite{Hager_2023_CVPR}. During the training phase, we fine-tune both the PENet and the Tabular Transformer by employing a combination of cross-entropy loss and CLIP loss. For the inference phase, we utilize the fine-tuned PENet and Tabular Transformer to perform predictions.
}
  \label{Clipfusion}
\end{figure}

We employed a multimodal fusion approach shown in Fig.~\ref{Clipfusion} inspired by the work of Hager et al.~\cite{Hager_2023_CVPR}, which itself is based on the principles derived from CLIP~\cite{radford2021learning}. We employed CLIP loss for fine-tuning our pretrained models, leveraging CLIP's exceptional performance not only in pre-training phases but also in fine-tuning scenarios, as highlighted by Dong et al.~\cite{dong2022clip}. The training and inference processes are outlined as follows:

\begin{itemize}
    \item Start with pre-trained models: PENet for PE images and a Tabular Transformer for EHR data.
    \item Proceed to fine-tune both the PENet and the Tabular Transformer using a hybrid loss function that combines cross-entropy and CLIP losses. The loss function is articulated as:
    \begin{equation}
    \begin{split}
    \mathcal{L} &= \lambda_1 \mathcal{L}_{\rm{c}}(z_{\mathrm{i}}, y) + \lambda_2 \mathcal{L}_{\rm{c}}(z_{\mathrm{t}}, y) \\ 
    &+ \lambda_3 \left(0.5 \mathcal{L}_{\rm{p}}(f_{\mathrm{i}}, f_{\mathrm{t}}) + 0.5 \mathcal{L}_{\rm{p}}(f_{\mathrm{t}}, f_{\mathrm{i}})\right),
    \end{split}
    \end{equation}
    where $\lambda_1 = 0.1$, $\lambda_2 = 0.1$, and $\lambda_3 = 0.8$ are the weighting coefficients. Here, $f_{\mathrm{i}}$ and $f_{\mathrm{t}}$ denote features extracted by the PENet and Tabular Transformer, respectively. $\mathcal{L}_{\rm{c}}$ represents the cross-entropy loss, and $\mathcal{L}_{\rm{p}}$ denotes the CLIP loss, which is defined as
    \begin{equation}
        \mathcal{L}_{\rm{p}}(f_{\mathrm{i}}, f_{\mathrm{t}}) = -\sum_{j \in N} \log \frac{\exp(\cos(f_{ji}, f_{jt}) / \tau)}{\sum_{k \in N, k \neq j} \exp(\cos(f_{ji}, f_{kt}) / \tau)},
    \end{equation}
    where $N = 8$ is the batch size and $\tau$ is set to 0.9995. The CLIP loss encourages the model to align images and tabular data features from the same case, while differentiating between features from different cases by minimizing or maximizing their cosine similarity.
    \item Conduct inference using the fine-tuned PENet and Tabular Transformer. We present three categories of results: (1) ensemble using outputs from both image and tabular data, (2) inference based on image data alone, and (3) inference based on tabular data alone.
\end{itemize}


Experimental results demonstrated in Table~\ref{Clip_comparison} indicate that single modality and ensemble surpassed the performance of CLIP fusion. The fact suggests that the efficacy of fusion methods is dependent on the datasets used. Some datasets contain multimodal data with high inter-modality correlation, which enhances the performance of fusion-based approaches. Conversely, in datasets where the multimodal data are relatively independent, fusion methods may not yield improvements.

\begin{table}[]
\resizebox{0.49\textwidth}{!}{\begin{tabular}{c|ccc|ccc}
\hline
       & \multicolumn{3}{c|}{CLIP fusion}   & \multicolumn{3}{c}{Single modality \& ensemble}       \\ \hline
       & Ensemble & Img only & Tab only & Ensemble & Img only & Tab only \\ \hline
AUPRC $\uparrow$ & 0.838    & 0.816      & 0.832      & \textbf{0.888}    & 0.848      & 0.842      \\
AUROC $\uparrow$ & 0.797    & 0.734      & 0.770      & \textbf{0.847}    & 0.786      & 0.788      \\
ACC $\uparrow$ & 0.705    & 0.663      & 0.621      & \textbf{0.774}    & 0.716      & 0.737      \\
MCC $\uparrow$ & 0.393    & 0.291      & 0.202      & \textbf{0.534}    & 0.407      & 0.472      \\
Recall $\uparrow$ & 0.918    & 0.882      & \textbf{0.964}      & 0.818    & 0.855      & 0.727      \\ \hline
\end{tabular}}
\caption{Performance comparison of models w/ and w/o CLIP fusion. CLIP fusion lead to a decline in most metrics.}
\label{Clip_comparison}
\end{table}

\section{SHAP-based Attribute Selection}

\begin{figure*}[tb]
  \centering
\includegraphics[angle=270,origin=c,width=0.40\textwidth]{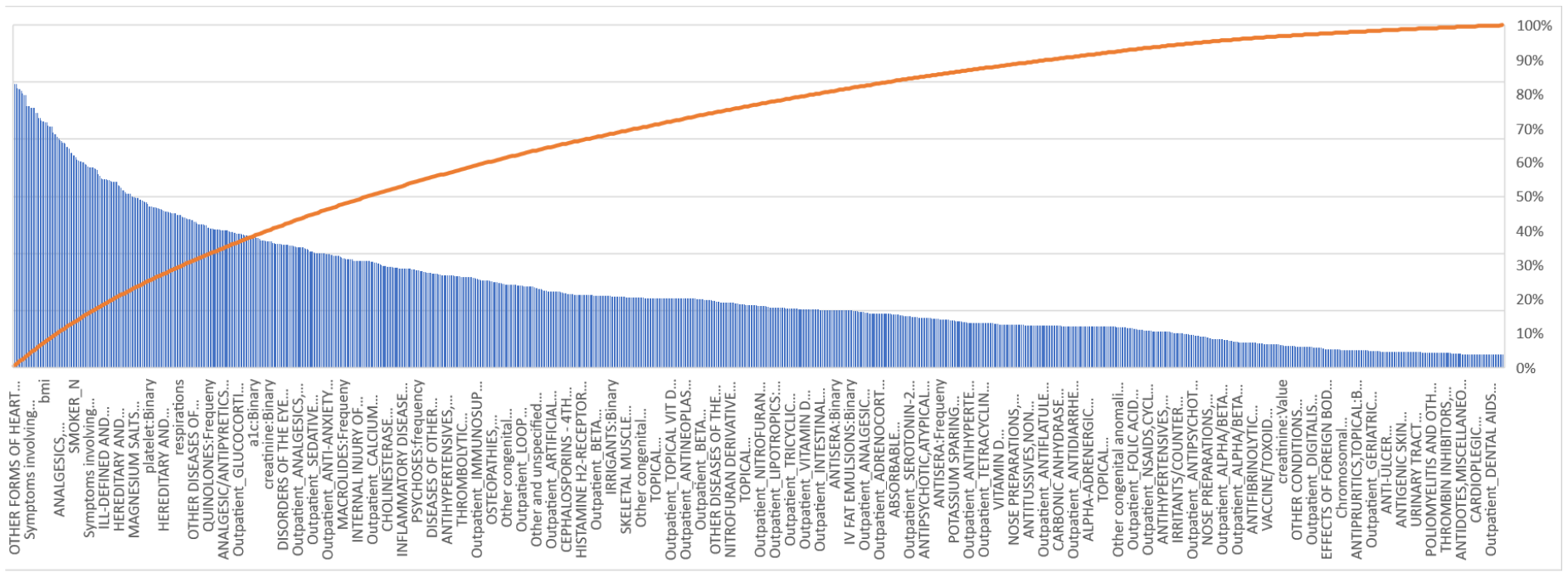}
  \caption{Top 250 influential attributes of the PE dataset's EHR tabular data.}
  \label{SHAP250}
\end{figure*}

\begin{figure*}[!htb]
  \centering
  \includegraphics[angle=90,origin=c,width=0.60\textwidth]{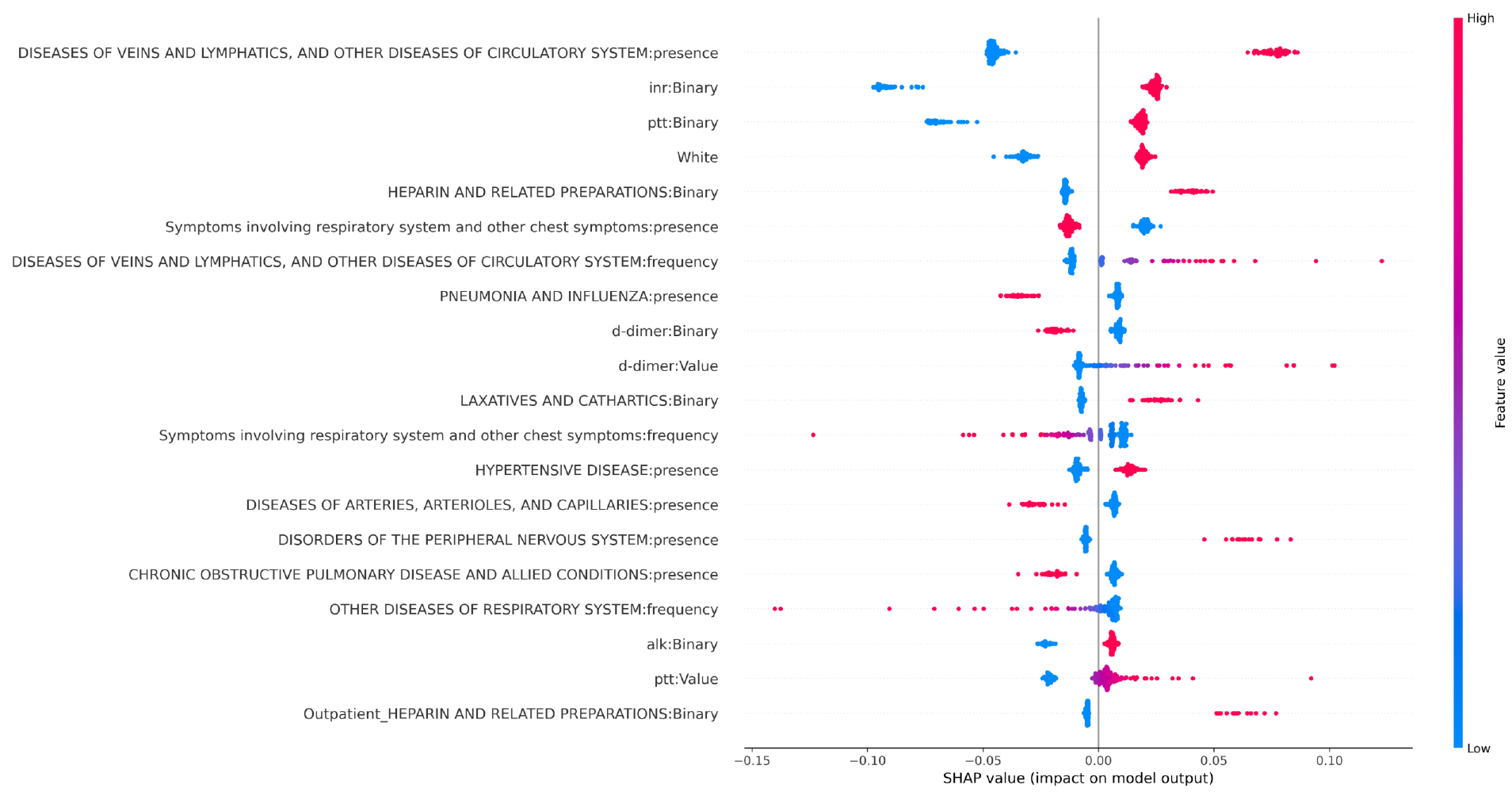}
  \caption{Closer examination of the top 20 attributes, along with their levels of contribution of the PE dataset's EHR tabular data.}
  \label{SHAP20}
\end{figure*}

\begin{table}[tb]
\centering
\resizebox{0.40\textwidth}{!}{\begin{tabular}{c|ccccc}
\hline
Attributes & 1503 & 1000  & 500   & 250   & 100   \\ \hline
AUROC     & 0.727 & 0.697 & 0.693 & \textbf{0.788} & 0.777 \\
ACC       & 0.694 & 0.705 & 0.684 & \textbf{0.737} & 0.737 \\
MCC       & 0.390 & 0.421 & 0.403 & \textbf{0.472} & 0.467 \\
\hline
\end{tabular}}
\caption{Impact of selecting different number of tabular attributes on the PE dataset}
\label{Tab_featureselect}
\end{table}

To enhance our classification model performance on the PE dataset, we applied SHAP (SHapley Additive exPlanations)~\cite{NIPS2017_7062} to identify the top 250 most impactful attributes from an initial set of 1,503 tabular attributes—down from 1,505 after excluding two directly related to PE. SHAP values, which are grounded in cooperative game theory, assign an importance score to each attribute based on its contribution to the model's prediction.

The formula to calculate SHAP values is given by:

\begin{equation}
\phi_i = \sum_{S} \frac{|S|! (|F| - |S| - 1)!}{|F|!} [f(S \cup \{i\}) - f(S)],
\end{equation}
where \(\phi_i\) is the SHAP value for feature \(i\), \(F\) is the total set of features, and \(S\) is a subset of features excluding \(i\).

The results, presented in Table~\ref{Tab_featureselect}, indicate that a subset of 250 attributes yielded superior classification accuracy compared to larger or smaller subsets of attributes. Illustrations of the top 250 influential attributes are provided in Fig.~\ref{SHAP250}, and a closer examination of the top 20 attributes, along with their levels of contribution, is depicted in Fig.~\ref{SHAP20}.

\section{Preprocessing of NACC Dataset}



\noindent \textbf{Case Selection and Image Preprocessing}
As detailed in the paper, the NACC dataset~\cite{beekly2004national} comprises 9,812 brain MRI cases featuring T1, T2, and FLAIR modalities. For the detection of cognitive impairment, the T1 modality is commonly preferred~\cite{zamani2022diagnosis}. The MRI cases selected for deep-learning classification adhered to the following criteria:
\begin{itemize}
    \item Inclusion of the T1 modality.
    \item Availability of high-resolution, thin-slice MRI scans.
    \item Accompanying tabular data encompassing cognitive assessments, gender, age, and other relevant information.
\end{itemize}
Following these criteria, we narrowed down the NACC dataset to 1,252 cases.

\begin{figure}[H]
  \centering
  \includegraphics[width=0.49\textwidth]{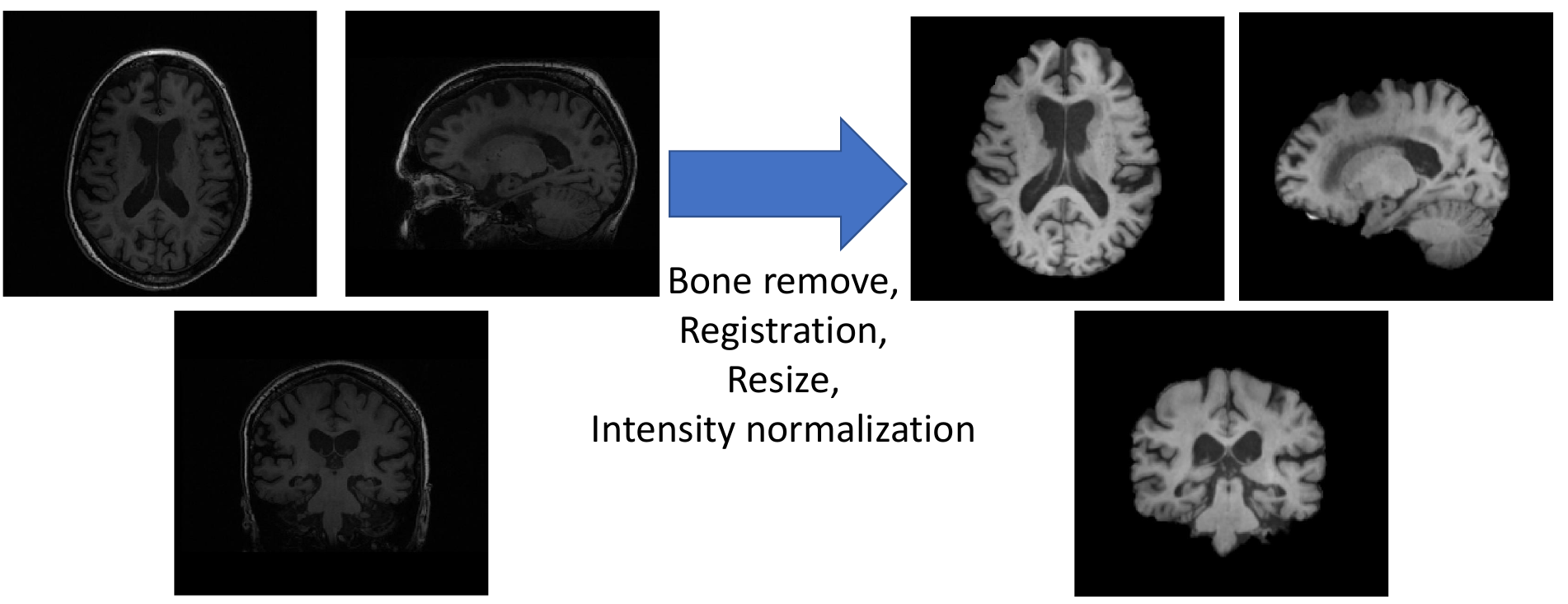}
  \caption{We employed FreeSurfer \protect\cite{fischl2012freesurfer} for skull stripping, resizing, and intensity normalization of NACC dataset's MRI volumes.}
  \label{NACC}
\end{figure}

Additionally, we employed FreeSurfer~\cite{fischl2012freesurfer} for skull stripping, resizing, and intensity normalization of the MRI data, as illustrated in Fig.~\ref{NACC}. After FreeSurfer's processing, each voxel is with dimensions of 182$\times$218$\times$182 voxels.

\section{Selecting Model for PE Tabular Classification}

ElasticNet~\cite{zou2005regularization}, known for combining ridge and lasso regression features, was previously used for tabular data classification in the PE dataset~\cite{zhou2021radfusion}. In our paper, the prediction is given by

\begin{equation}
\hat{y} = f_{\rm{a}}(\boldsymbol{\beta}^\top \boldsymbol{x}_{\rm{t}} + \beta_0)
\end{equation}
where each tabular data vector is denoted as a feature vector $\boldsymbol{x}_{\rm {t}}$ and the objective is to predict the class label by output $\hat{y}$. $\boldsymbol{\beta}$ is the vector of coefficients and $\beta_0$ is the intercept. The function $f_{\rm{a}}(\cdot)$ represents the activation function, mapping the linear combination to a class probability.

The coefficients $\boldsymbol{\beta}$ and $\beta_{0}$ are derived via ElasticNet regularization:

\begin{equation}
\min_{\boldsymbol{\beta}} \frac{1}{N} \sum_{n=1}^{N} \mathcal{L}(y_n, \hat{y}_n) + \lambda \left[(1 - \alpha) \frac{1}{2} \|\boldsymbol{\beta}\|_2^2 + \alpha \|\boldsymbol{\beta}\|_1\right],
\end{equation}
where $\mathcal{L}$ is the loss function, $N$ is the number of instances, $\lambda$ the regularization weight, and $\alpha$ balances L1 and L2 penalties, facilitating feature selection and robustness in high-dimensional spaces.

ElasticNet demonstrated good performance (ACC=0.837) as reported in Zhou et al.~\cite{zhou2021radfusion} when all 1,505 attributes of the PE dataset's tabular data were utilized. As stated in the paper, this high accuracy can be attributed to the inclusion of two attributes directly related to PE: ``DISEASES OF PULMONARY CIRCULATION: frequency" and ``DISEASES OF PULMONARY CIRCULATION: presence." Our analysis revealed that if these two attributes are not 0, PE is positive. Consequently, to avoid bias, we decided to exclude these two attributes from our analysis.

Given the advancements in machine learning since ElasticNet's introduction in 2005, we explored newer models to evaluate their effectiveness. This included the deep learning-based ResNet~\cite{jian2016deep}, known for its deep network capabilities through skip connections; and the Tabular Transformer~\cite{NEURIPS2021_9d86d83f}, which adapts the transformer architecture for structured data.

Our analysis shown in Table 5 in the paper, indicated that the Tabular Transformer outperformed other models, including ElasticNet and ResNet, in classifying the PE dataset. Consequently, we chose the Tabular Transformer as our baseline model due to its superior performance in handling complex tabular data.

\section{Additional Grad-CAM Results}

\begin{figure*}[tb]
  \centering
  \includegraphics[angle=90,origin=c,width=0.70\textwidth]{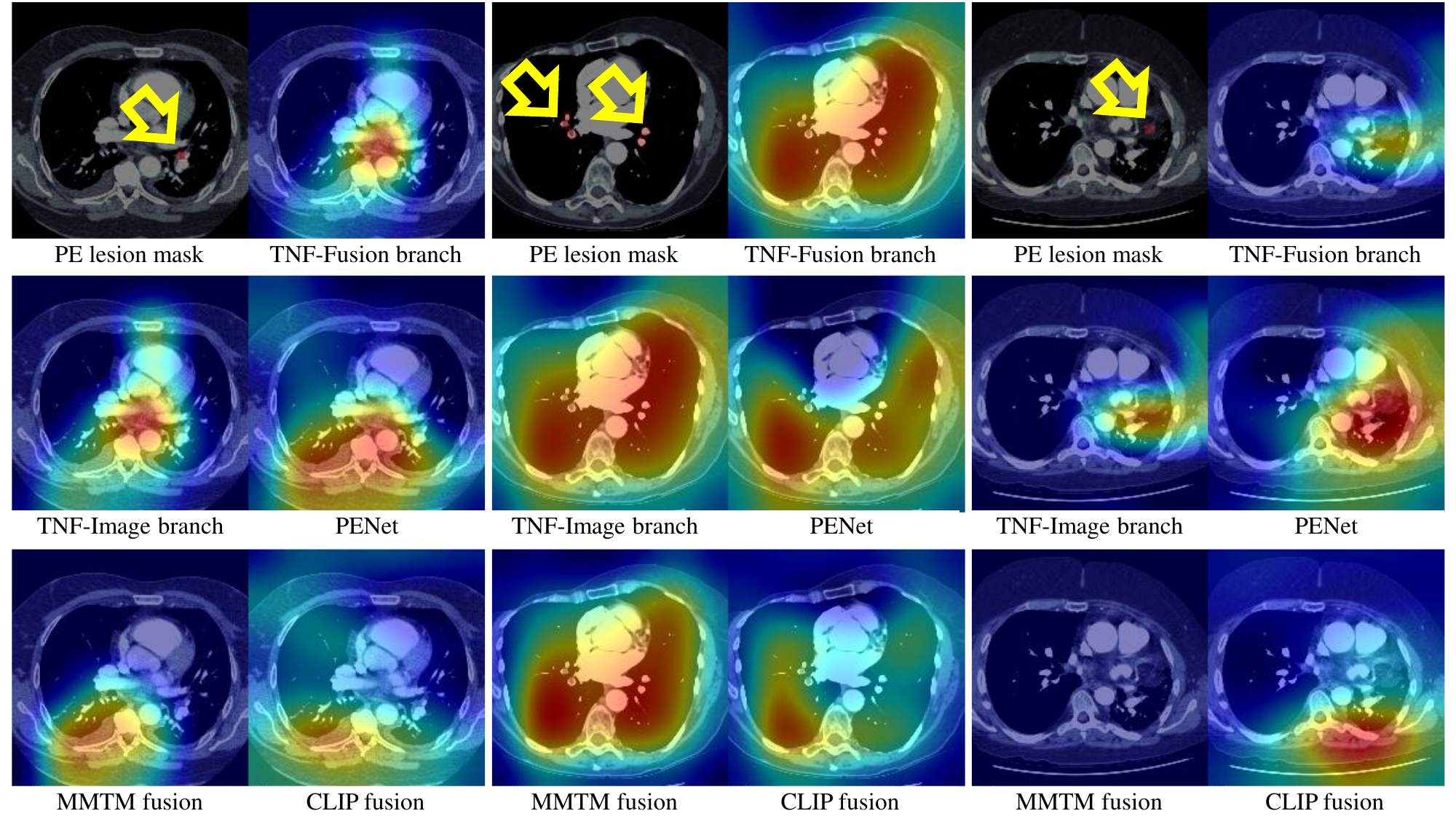}
  \caption{More Grad-CAM heatmaps. Yellow arrows indicate PE lesions.}
  \label{Appendix_GradCAM}
\end{figure*}

To demonstrate the superior performance of TNF-based models compared to others, additional Grad-CAM heatmaps are presented in Fig.~\ref{Appendix_GradCAM}. We compared five types of Grad-CAM heatmaps: 1) The fusion branch of TNF. The model is shown in Fig.~\ref{PE_Nets} (a); 2) The image branch of TNF; 3) PENet; 4) MMTM fusion model; The model's structure is equivalent to the model in Fig.~\ref{PE_Nets} (a) and only outputs fusion result; 5) Clip fusion model. Yellow arrows in Fig.~\ref{Appendix_GradCAM} indicate PE lesions. In short, the highlighted parts of the TNF-based models are closer to the lesions.

\newpage

\bibliographystyle{named}
\bibliography{ijcai23}

\end{document}